\documentclass[letterpaper, 10pt, conference]{ieeeconf}
\IEEEoverridecommandlockouts
\usepackage{cite}
\usepackage{amsmath,amssymb,amsfonts}
\usepackage{algorithmic}
\usepackage{graphicx}
\usepackage{textcomp}
\usepackage{xcolor}
\usepackage{tikz}
\usetikzlibrary{positioning,arrows.meta,fit,backgrounds,calc}
\usepackage{booktabs}      
\usepackage{siunitx}       
\usepackage[hidelinks]{hyperref}
\usepackage{cleveref}
\usepackage{url}

\usepackage[normalem]{ulem}

\newcommand{\rev}[1]{#1}

\def\BibTeX{{\rm B\kern-.05em{\sc i\kern-.025em b}\kern-.08em
    T\kern-.1667em\lower.7ex\hbox{E}\kern-.125emX}}

\newif\ifblind
\blindfalse
\crefname{figure}{Fig.}{Figures}
\crefname{section}{Sec.}{Sections}
\crefname{equation}{Eq.}{Equations}
\crefname{table}{Table}{Tables}
\begin{document}

\title{Personalization of \rev{a} Commercial Powered Knee Prosthesis 
Under Clinically Realistic Conditions via Reinforcement
Learning: A Feasibility Study
}

\ifblind
  \author{Anonymous Authors}
\else
  \author{Xinyi Liu$^{1}$, Atli Örn Sverrisson$^{2}$, David Langlois$^{2}$, 
    Ming Liu$^{1}$, 
    Varun Nalam$^{1}$, and He Huang$^{1}$%
    \thanks{  This material is based upon work supported by the Eunice Kennedy Shriver National Institute of Child Health and Human Development of the National Institutes of Health under Award R01HD116422. Approximately 100\% of the total project costs were financed with federal funds. 
The content is solely the responsibility of the authors and does not necessarily represent the official views of the National Institutes of Health. 
   }%
    \thanks{$^{1}$Lampe Joint Department of Biomedical Engineering,
      University of North Carolina at Chapel Hill and North Carolina State
      University, Raleigh, NC, USA.}%
    \thanks{$^{2}$\"Ossur, Reykjav\'{\i}k, Iceland.}%
  }
\fi

\maketitle
\thispagestyle{empty}
\pagestyle{empty}

\begin{abstract}
Personalization of powered prosthesis control to individual user needs remains a significant challenge. To address this challenge{,} policy iteration with constraint embedded (PICE), {a reinforcement learning (RL)} tuning approach{,} has been developed and validated. This feasibility study investigated whether the developed approach can personalize a commercial powered knee prosthesis under clinically realistic conditions. To reduce the time for personalization, {tuning} was initialized with a policy learned offline from one non-disabled participant walking on a treadmill and was then applied to three other non-disabled participants walking with the knee through a bypass adapter on a 5-meter walkway. All participants reached the goal within 8--18 tuning steps (1.7--4.2~minutes of walking), reducing the knee-kinematic feature error by 64--90\% on average. These results support the feasibility of implementing the RL-based tuning approach in clinically realistic conditions with space and equipment constraints.
\end{abstract}



\section{Introduction}

For individuals with transfemoral amputation, powered knee prostheses can generate positive mechanical power during gait, which passive knees cannot~\cite{tucker_control_2015}. Research prototypes have demonstrated clinical benefits in laboratory settings~\cite{lawson_robotic_2014}, 
and clinical evaluations of commercial devices have reported improved biomechanical outcomes and favorable user feedback~{\cite{best_clinical_2025}}, yet these benefits are not fully realized across {users}~{\cite{hafner_physical_2015, kestur_comparing_2024, herrin_toward_2026}}. This inconsistency has limited the adoption of powered prostheses in clinical practice and everyday use~{\cite{best_clinical_2025}}.
{A major bottleneck is personalizing the controller to each user~\cite{wen_online_2020, slade_human---loop_2024}.} {Across} robotic lower-limb prostheses, user outcomes depend on how the controller is configured~\cite{best_clinical_2025, gehlhar_review_2023, rogers-bradley_variable-stiffness_2024}. {Commonly used prosthesis control approaches, from} finite-state impedance control~\cite{simon_configuring_2014, lawson_robotic_2014}, continuous impedance control~\cite{hong_offline_2025}, hybrid kinematic--impedance control~\cite{reznick_clinical_2025}, and model-based control~\cite{zhao_first_2017} to recent learning-based controllers trained in simulation~\cite{le_replay-constrained_2026}\rev{,} all require per-user customization to achieve clinical benefits~\cite{gehlhar_review_2023}.

Personalization must account for {many user-specific factors and is predominantly performed manually without a standardized procedure. Parameters must be adjusted to each user's body size and gait style~\cite{ichimura_spatiotemporal_2022, kestur_comparing_2024}.
 {Some factors} also {change over time}: 
users re-weight movement goals such as efficiency and stability as environmental risk changes~\cite{kulkarni_humans_2023, feldman_estimating_2026}, and users adapt their motor control to the device~\cite{huang_taking_2021}{, so a} configuration tuned once may need to be revisited. In current practice, prosthetists tune {the parameters} by observing the user across walking trials~\cite{simon_configuring_2014}{, which is} time-consuming and relies on expert knowledge of many interacting parameters: even with a structured starting configuration, 24 of the 140 parameters of a powered knee--ankle prosthesis had to be adjusted for each user, taking up to 5~hours in one such approach~\cite{simon_configuring_2014}. Moreover, clinical evaluations often do not report the tuning procedure or criteria used \cite{best_clinical_2025, kestur_comparing_2024}, which makes outcomes difficult to compare across studies.

\begin{figure}[t]
    \centering
    \includegraphics[width=\linewidth]{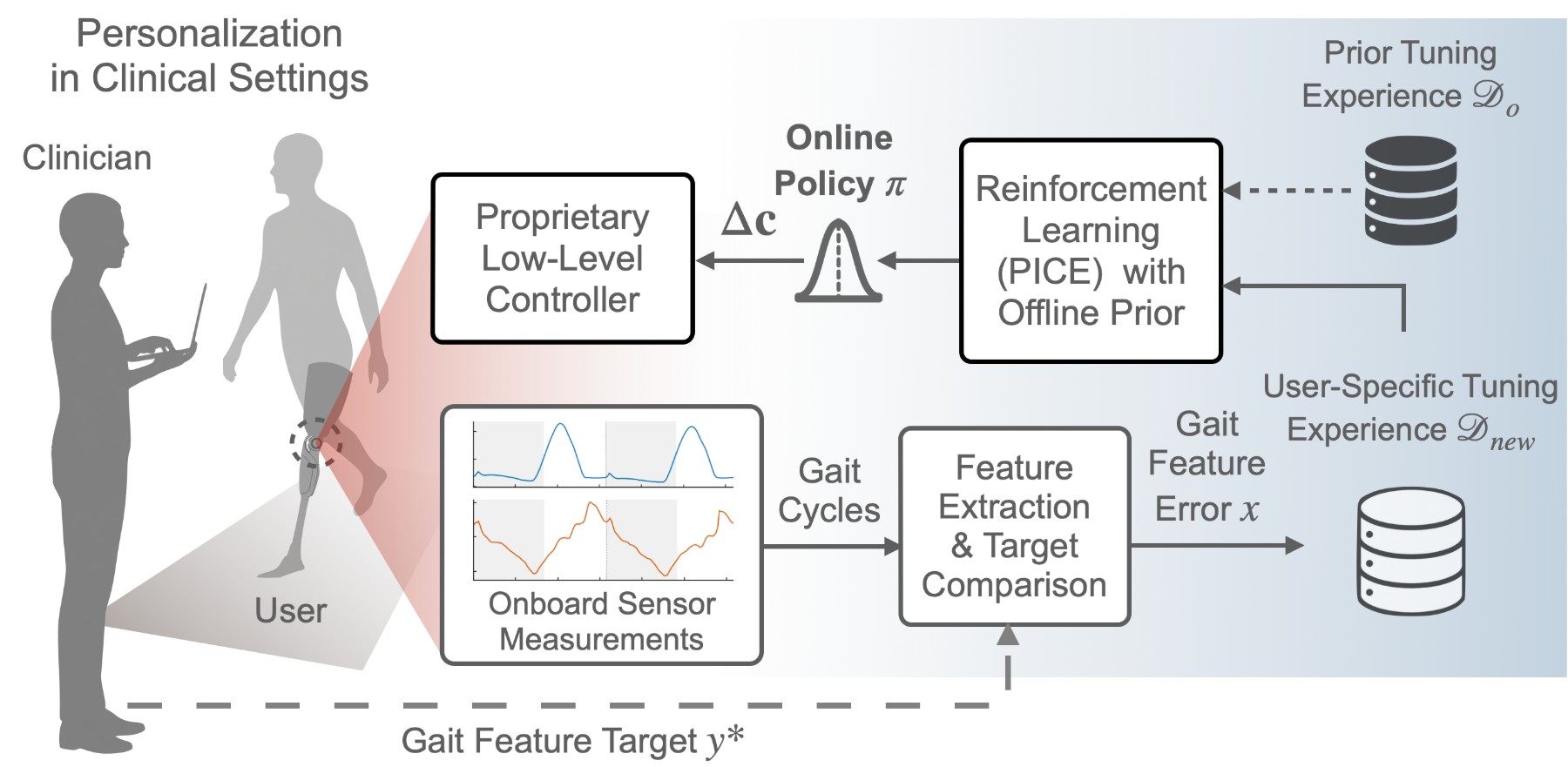}
    \caption{Conceptual overview of prior-informed prosthesis personalization in an envisioned clinical workflow. A clinician specifies a target set of gait features, and sensor measurements from the prosthetic knee are used to quantify deviations from the prescribed target. PICE initializes a tuning policy $\pi$ using prior experience $\mathcal{D}_o$ and updates the policy using user-specific tuning experience $\mathcal{D}_{\mathrm{new}}$ collected during walking. The updated policy generates parameter adjustments $\Delta\mathbf{c}$ for the proprietary low-level controller, enabling iterative personalization through a closed-loop tuning process without replacing the underlying controller. Red and blue indicate onboard and offboard computation, respectively.}
    \label{fig:pice}
\end{figure}

Automating tuning toward a prescribed goal could shorten clinic visits, standardize fitting procedures, reduce dependency on subjective expert tuning, which could be biased, and let users benefit more fully from powered prostheses. Such automation is commonly framed as human-in-the-loop optimization (HILO)~\cite{slade_human---loop_2024}. Black-box optimizers that search for a single best parameter set, such as covariance matrix adaptation evolution strategy and {Bayesian} optimization,  have reduced the metabolic cost of walking with exoskeletons~\cite{zhang_human---loop_2017, ding_human---loop_2018}, but \rev{have not outperformed} generic control for prosthesis users~\cite{welker_shortcomings_nodate}, likely because users' motor control and objectives change as they adapt to the device~\cite{welker_shortcomings_nodate, huang_taking_2021}. Another common HILO framework is reinforcement learning (RL) with human-subject testing in the loop, which instead drives the prosthesis motion toward a target profile~\cite{fylstra_human-prosthesis_2022, le_replay-constrained_2026}. Rather than searching for a single optimum, it learns a \emph{tuning policy} that maps the observed knee-motion error to incremental parameter adjustments~\cite{wen_online_2020}, which can be reapplied whenever the user's condition changes. Policy iteration with constraint embedded (PICE)~\cite{li_toward_2022} tuned 12 impedance parameters within 45 tuning steps, and the resulting policies remained effective across users and task modifications. RL tuning has since been extended to continuous impedance control~\cite{hong_offline_2025} and to shaping the user's thigh motion~\cite{liu_addressing_2026}. RL tuning frameworks such as PICE could therefore serve as a general tuning mechanism, but there are still challenges to be addressed to ensure practical implementation in clinics~\cite{wen_online_2020, li_toward_2022, hong_offline_2025}.

Our previously validated PICE algorithm~\cite{li_toward_2022} has so far been used to tune prosthesis impedance only during treadmill walking, where continuous walking at a fixed speed keeps gait variability low~\cite{hollman2016comparison}. In clinical practice, however, prostheses are typically tuned during self-paced overground walking~\cite{reznick_clinical_2025}, which produces more variable gait and is interrupted by frequent stops and turns imposed by limited walkway space. Whether PICE can tune a prosthesis effectively under these clinically realistic conditions remains an open question.
In this study, we tested PICE on a commercial powered knee prosthesis under such conditions. We further asked whether tuning for a new user can be {accelerated by} reusing prior experience from another user. Building on~\cite{li_toward_2022}, which showed that offline data can improve tuning efficiency, we initialized tuning with a policy learned offline from one non-disabled participant walking on a treadmill, rather than starting from scratch. We then used this policy to tune the knee for three other non-disabled participants walking overground with a bypass adapter, {and, for one participant, toward a second set of target knee kinematics}. We show that all participants reached the goal within 8--18 tuning steps, {corresponding to} 1.7--4.2~minutes of walking, reducing the knee-kinematic feature errors by 64--90\%. These results \rev{are a preliminary step toward clinically relevant automated prosthesis tuning}.

The remainder of this paper is organized as follows. 
Section~\ref{sec:formulation} {describes the system setup and} formulates the problem,
Section~\ref{sec:methods} describes {the tuning method and evaluation protocol,} Section~\ref{sec:results} reports results, and
{Section}~\ref{sec:discussion} {discusses} limitations
and {concludes}.

\section{System Setup \& Problem  Formulation}
\label{sec:formulation}
\subsection{System Setup}
\begin{figure}[t]
    \centering
    \includegraphics[width=\linewidth]{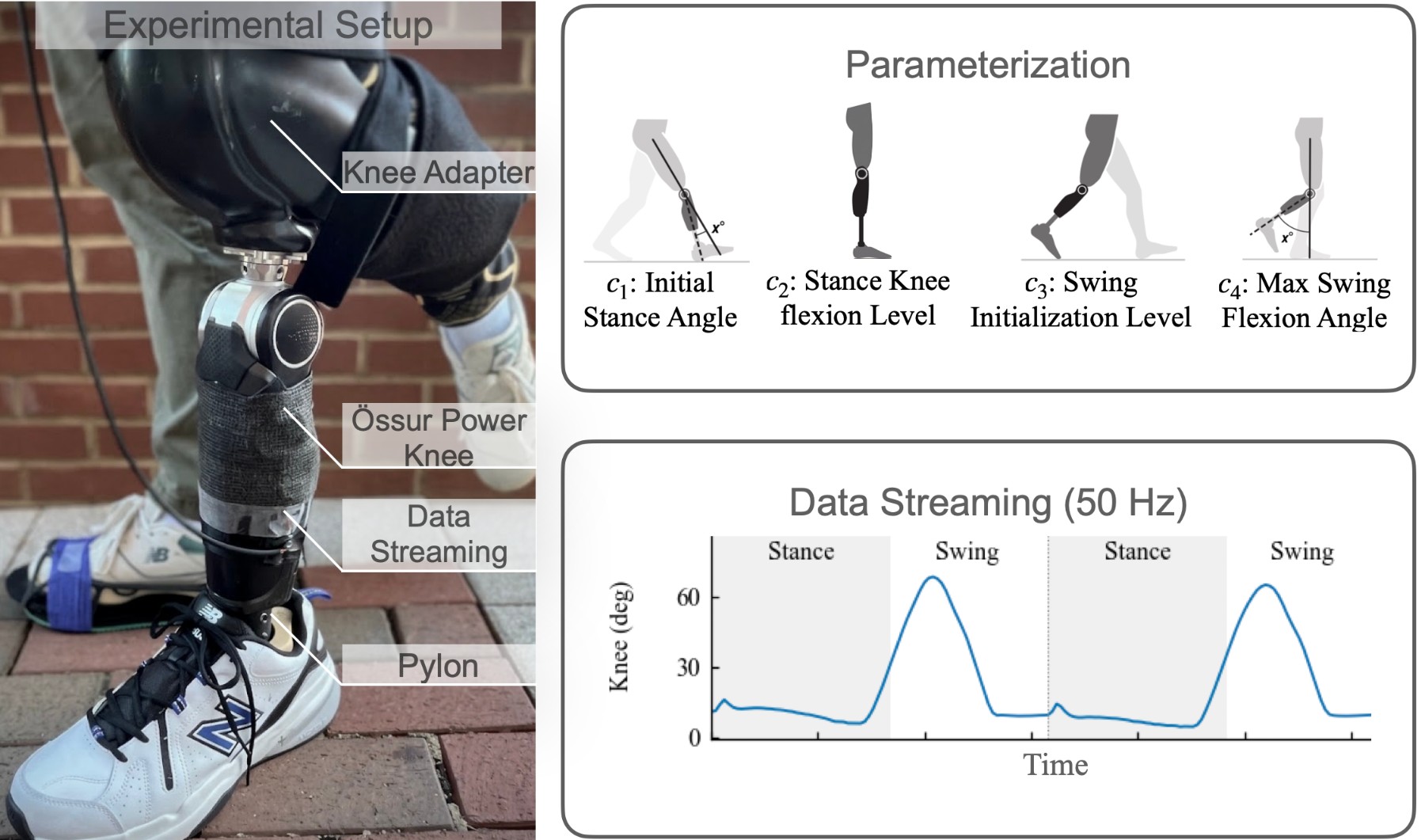}
    \caption{Experimental setup and parameterization. Left: participant wearing
    the {\"Ossur} Power Knee, with a wire streaming data from the knee. Top right:
    the four clinical parameters tuned by the policy, $c_1$ initial stance
    angle, $c_2$ stance knee flexion level, $c_3$ swing initialization level,
    and $c_4$ maximum swing flexion angle. Bottom right: knee angle
    over two gait cycles, streamed from the knee's onboard sensors at {about} 50~Hz; shaded regions mark stance.}
    \label{fig:setup}
\end{figure}

The tuning was performed on the Power Knee from Össur\footnote{Model version PKA01, firmware version PKK3MMD0302062}, a \rev{commercially available} robotic knee prosthesis, with non-disabled participants. The prosthesis was mounted on an
L-shaped bypass adapter and connected distally to a pylon and \rev{an} unactuated
prosthetic foot (\cref{fig:setup}, left). The knee's proprietary
low-level controller was {treated as a black box}. 
Tuning used only the knee's onboard sensors, with the policy computed on a laptop connected to the knee{;} no motion capture or other laboratory equipment was
needed. 
The tuner interacted with the knee only
through four clinical parameters exposed for prosthetist fitting
(\cref{fig:setup}, top right): the initial stance angle $c_1$, the
stance knee flexion level $c_2$, the swing initialization level $c_3$, and
the maximum swing flexion angle $c_4$, each normalized to $[0,1]$. \rev{Knee angles} measured by the knee's onboard sensors
were streamed at about 50~Hz (\cref{fig:setup}, bottom right) to
a laptop, which received the stance and swing phases segmented by the
knee's proprietary firmware from its load-cell measurements, extracted the gait features, and sent
the updated parameters back to the knee.

\subsection{Problem Formulation}
We formulate clinical tuning as {an RL problem.}
The unknown dynamics of the \rev{human--robot} coupled system and the low-level control are treated as a black box whose only interface is the vector of four clinical
parameters $\mathbf{c}=[c_1,\dots,c_4]^\top$  (\cref{fig:setup}), bounded by the normalized set
$\mathcal{C}=[0,1]^4$. 
We selected phase-specific knee-kinematic extrema as tuning
features:  the peak flexion and peak
extension angles in stance and in swing,
$\mathbf{y}=[\theta^{\mathrm{st}}_{\max},\theta^{\mathrm{st}}_{\min},
\theta^{\mathrm{sw}}_{\max},\theta^{\mathrm{sw}}_{\min}]^\top$, with 
ranges $[\underline{y}_i,\overline{y}_i]$ of $[4,30]$, $[0,18]$, $[40,80]$,
and $[4,13]$~deg, respectively. These features were selected because they directly characterize the powered prosthesis knee motion influenced by the clinical parameters and can be reliably extracted from onboard measurements for online tuning. The feature ranges were set from the variability in previously collected gait data and limited to values considered safe for walking.
Compared with broader gait outcomes, such as symmetry, joint kinetics, or metabolic cost, they provide a compact, phase-specific, and clinically interpretable objective without requiring additional instrumentation or full-trajectory comparison \cite{alili2023novel}. 


Tuning occurred every 6 gait cycles, and at each tuning step $k$,  
knee-kinematic features were extracted from the average knee profile of the {most} recent 4 gait cycles. The state
$\mathbf{x}_k=[x_1,\dots,x_4]^\top$ is the error of each feature from its
target $y_i^\ast$, normalized by \rev{half} the width of its range, \(
  x_i = 2(y_i - y_i^\ast)/(\overline{y}_i - \underline{y}_i),
   i = 1,\dots,4 .
\)
This formulation permits user-specific target selection within the prescribed feature ranges. The policy maps the state to a parameter increment as the action,
$\mathbf{a}_k=\pi(\mathbf{x}_k)$, and the parameters are updated as
$\mathbf{c}_{k+1}=\Pi_{\mathcal{C}}(\mathbf{c}_k+\mathbf{a}_k)$,
where $\Pi_{\mathcal{C}}$ projects onto $\mathcal{C}$. Each tuning step yields the reward
\begin{equation}
  r_k = -\left(\mathbf{x}_{k+1}^\top\mathbf{W}\,\mathbf{x}_{k+1}
        + \mathbf{a}_k^\top\mathbf{R}\,\mathbf{a}_k\right),
  \label{eq:cost}
\end{equation}
which penalizes the weighted feature error and the size of each adjustment,
with $\mathbf{W} = I$ and $\mathbf{R} = 0.3 I$.

The goal is to train a policy
that maximizes the action-value function
\begin{equation}
  Q^{\pi}(\mathbf{x},\mathbf{a})
  = \sum_{k=0}^{\infty}\gamma^{k} r_k ,
  \quad
  \mathbf{a}_k=\pi(\mathbf{x}_k),
  \label{eq:qvalue}
\end{equation}
i.e., $\pi^\ast=\arg\max_{\pi} Q^{\pi}$, where $\gamma = 0.95$ is the
discount factor. The
transition from $\mathbf{x}_k$ to $\mathbf{x}_{k+1}$ depends jointly on the
black-box controller and the user, \rev{neither of which is modeled}; $\pi$ is learned
model-free with PICE~\cite{li_toward_2022} \rev{(Section~\ref{sec:methods})}.

\section{Methodology}
\label{sec:methods}

\subsection{Policy Iteration with Constraint Embedded}
\label{subsec:pice}
We adopted PICE~\cite{li_toward_2022}, a constrained variant of least-squares temporal-difference Q-learning (LSTD-Q)~\cite{lagoudakis_least-squares_2003}, as the {RL} backbone. PICE is model-free, data-efficient, and has been previously validated for prosthesis tuning~\cite{li_toward_2022,hong_offline_2025}. It approximates the action-value function $\hat{Q}^{\mathbf{w}}(\mathbf{x},\mathbf{a})$ using a parameterized function with weights $\mathbf{w}$.
The resulting tuning policy selects the action that maximizes the estimated action-value function. Specifically, the negative action-value function is expressed as a quadratic cost $\mathbf{z}^{\top}\mathbf{H}\mathbf{z}$, where $\mathbf{z} = [\mathbf{x}^{\top},\mathbf{a}^{\top}]^{\top}$ is the augmented state-action feature vector and $\mathbf{H}$ is a symmetric matrix uniquely determined by $\mathbf{w}$, since $\mathbf{w}$ collects the entries of the upper-triangular part of $\mathbf{H}$. At each tuning step, the policy selects the parameter increment {$\mathbf{a}=\Delta\mathbf{c}$} within the admissible action space $\mathcal{A}=[-0.15,0.15]^4$ by solving
\begin{equation}
  \pi(\mathbf{x}) = \arg\max_{\mathbf{a}\in\mathcal{A}}\hat{Q}^{\mathbf{w}}(\mathbf{x},\mathbf{a})
  = \arg\min_{\mathbf{a}\in\mathcal{A}}\;
  \mathbf{z}^\top\mathbf{H}\,\mathbf{z}\rev{.}
  \label{eq:qp}
\end{equation}
\rev{PICE} constrains the learned weights to the feasible set
$
\mathcal{E}=
\{
\mathbf{w}:
\mathbf{H}\succeq 0,\ \|\mathbf{w}\|\le {\rho}
\},
$
where $\mathbf{H}\succeq 0$ denotes positive semidefiniteness and {$\rho=100$} bounds the weight {norm}. These constraints ensure that the resulting optimization problem is convex and admits a globally optimal solution.

Initializing the learning process from scratch may require substantial clinical tuning time~\cite {li_toward_2022}. To reduce this burden, we constructed an offline prior using previously collected prosthesis tuning data. Specifically, an offline dataset $\mathcal{D}_o$ consisting of $N_o=150$ tuning steps was collected from one non-disabled participant (AB0) walking on a treadmill at 0.8~m/s. During data collection, {the clinical} parameters were adjusted using randomly sampled increments from $\mathcal{A}$ to explore the admissible action space.
PICE was then applied to $\mathcal{D}_o$ to obtain the offline policy weights $\mathbf{w}_o$. 
For each new user, the tuning process was initialized with the offline policy parameterized by $\mathbf{w}_o$. 
During online tuning, each tuning step added one sample $(\mathbf{x}_k,\mathbf{a}_k,r_k,\mathbf{x}_{k+1})$ to a user-specific dataset $\mathcal{D}_{new}$. Every $N_b=12$ tuning steps, PICE was rerun to update {the} learned weights $\mathbf{w}$ following the procedure in \cite{li_toward_2022}.

\subsection{Participants and Protocol}
\begin{table}[t]
\centering
\caption{Study participant information and experimental design.}
\label{tab:design}
\setlength{\tabcolsep}{4.8pt}
\begin{tabular}{lllccccc}
\midrule
 & & & Age & Height & Mass & \multicolumn{2}{c}{Target $y^*$}\\
\cmidrule(lr){7-8}
ID & Role & Sex & (yr) & (m) & (kg) & T1 & T2\\
\midrule
AB0 & Prior  & M & 30 & 1.73 & 75.0 &            &           \\
AB1 & Tuning & M & 21 & 1.70 & 68.0 & \checkmark &           \\
AB2 & Tuning & M & 30 & 1.76 & 86.2 & \checkmark & \checkmark\\
AB3 & Tuning & M & 33 & 1.70 & 70.3 & \checkmark &        \\
\midrule
\end{tabular}
\end{table}

 The study was approved by the Institutional Review Board of the University of
North Carolina at Chapel Hill (protocol \#29310), and all participants gave
written informed consent.
All completed trials were analyzed regardless of outcome.

Participants walked at a self-selected speed on a 5-meter level walkway with an overhead safety harness. Due to limited walking distance, gait cycles associated with stopping, turning, and gait initiation were excluded from feature extraction and tuning updates to avoid transient gait patterns.
The algorithm tuned knee motion {toward} specified targets only during the {steady-state gait cycles}. To evaluate the feasibility of this framework under controlled experimental conditions, the gait feature targets were predefined and held constant within each trial. {Target T1~$= [12.0, 8.0, 63.0, 9.0]$~deg, derived from previously collected gait data, was used for all participants. A second target, T2~$= [8.0, 2.0, 50.0, 8.0]$~deg,}
was additionally tested {with AB2} in a {separate} trial to evaluate adaptation to different targets without modifying the optimization procedure. The target order for AB2 was randomized. Each trial started from the prior policy (\rev{Section~\ref{subsec:pice}}) and a random initial parameter setting $\mathbf{c}_0$.

A trial was considered successful if, within 10 minutes of walking, both criteria were satisfied in at least four of the five most recent tuning steps: (i) all normalized feature errors satisfied $|x_i| < 0.3$, corresponding to 3.9, 2.7, 6.0, and 1.4 deg for the stance maximum, stance minimum, swing maximum, and swing minimum, respectively; and (ii) the reward (\cref{eq:cost}) exceeded $-0.5$. Otherwise, the trial was terminated and classified as a failure. 
The success criteria were defined to evaluate convergence toward the prescribed kinematic targets rather than to establish clinically optimal gait performance. The thresholds were selected based on prior observations of kinematic variability during self-paced walking.

\subsection{Outcome Measures and Statistics}
The primary outcomes were {trial success or failure}, the absolute {feature} errors
{$|y_i - y_i^*|$} at the first and last tuning steps, and the number of tuning steps,
with the corresponding net walking time, defined as the total trial time minus the time during transient steps. As secondary outcomes, we report the reward $r_k$ in {\cref{eq:cost}} at each tuning step and the knee-angle profile over the
gait cycle at the start and end of tuning.
When a trial triggered an online policy update, we evaluated the update with the temporal-difference (TD) error of the learned action-value function on the transitions collected after the update,
\begin{equation}
\delta_k = r_k + \gamma\,\hat{Q}^{\mathbf{w}}\big(\mathbf{x}_{k+1},\pi(\mathbf{x}_{k+1})\big) - \hat{Q}^{\mathbf{w}}(\mathbf{x}_k,\mathbf{a}_k),
\label{eq:td}
\end{equation}
where $\pi$ is the greedy policy of \cref{eq:qp}. The TD error is the residual of the Bellman equation: it is zero when $\hat{Q}^{\mathbf{w}}$ exactly predicts the observed reward plus the discounted value of the next state. We report its root-mean-square (RMS) value, \rev{$\sqrt{\tfrac{1}{n}\sum_{k=1}^{n} \delta_k^2}$, where $n$ is the number of evaluated transitions}, for the offline policy and the updated policy, both evaluated on tuning steps that neither policy was trained on.
With {one to three} participants per condition
(Table~\ref{tab:design}), results are reported per participant and
condition, without statistical tests.

\section{Results}
\label{sec:results}

\begin{figure}
    \centering
    \includegraphics[width=1\linewidth]{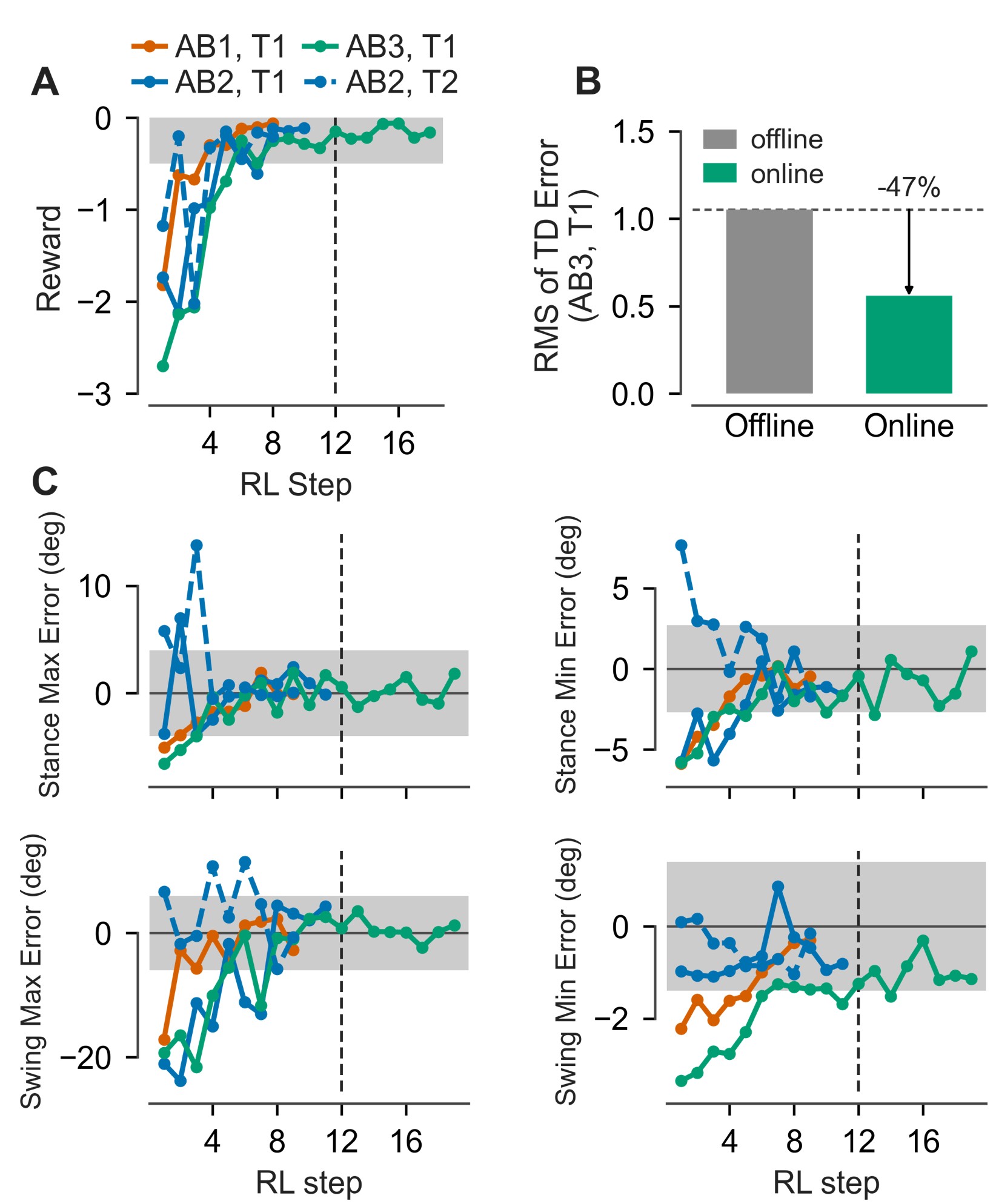}
    \caption{Online tuning in the four trials: AB1, AB2, and AB3 with target T1 (solid lines) and AB2 with target T2 (dashed line). (A)~Reward at each tuning (RL) step. (B)~\rev{Root-mean-square} (RMS) value of \rev{temporal-difference} (TD) error (\cref{eq:td}) of AB3's learned action-value function on the held-out steps 13--18, for the offline policy and the policy after the first online update (online). (C)~Error of each knee-kinematic feature from its target, $y_i - y_i^*$, in degrees. Gray bands mark the goal range: a reward above $-0.5$ in (A) and the goal threshold of each feature in (C). Vertical dashed lines in (A) and (C) mark step~12, the first scheduled online policy update ($N_b = 12$), which only AB3 reached.}
    \label{fig:result_main}
\end{figure}

\begin{figure}
    \includegraphics[width=1\linewidth]{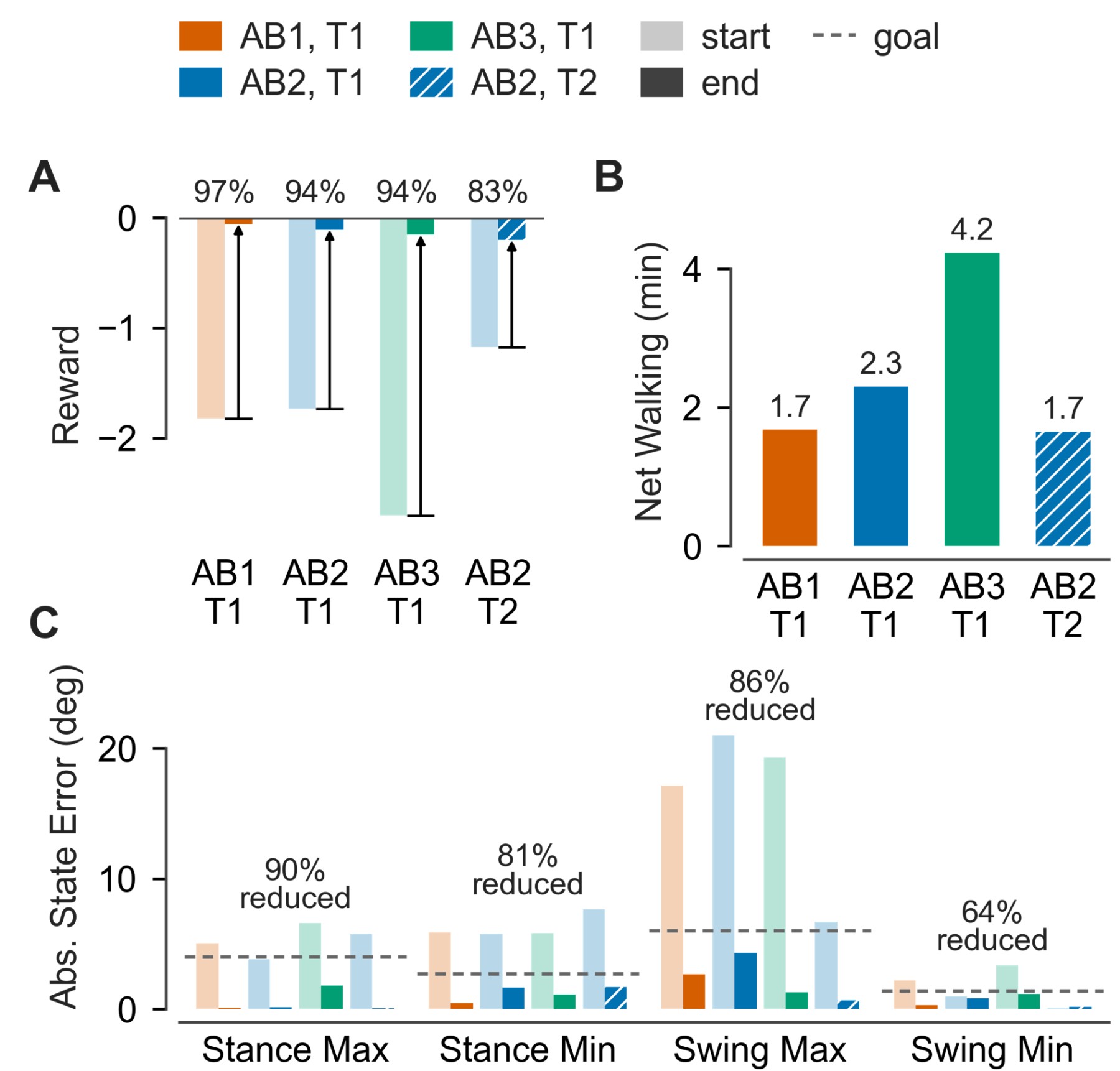}
    \caption{Start versus end of tuning for the four trials. (A)~Reward at the first (light) and last (dark) tuning step; arrows show the change and percentages the improvement. (B)~Net walking time until the success criterion was met. (C)~Absolute error of each knee-kinematic feature, $|y_i - y_i^*|$, at the first (light) and last (dark) tuning step; dashed lines mark the goal threshold of each feature, and percentages give the reduction in the error averaged over the four trials. Hatched bars: AB2 with target T2.}
    \label{fig:stats}
\end{figure}


Fig.~\ref{fig:stats} summarizes every trial. All four completed trials met the success criterion within 8--18 RL steps (1.7--4.2~min of walking\rev{;} Figs.~\ref{fig:result_main}A and~\ref{fig:stats}B), well within the 10-minute limit. Averaged over the four trials, the absolute error fell by 90\%, 81\%, 86\%, and 64\% for the stance maximum, stance minimum, swing maximum, and swing minimum, respectively (\cref{fig:stats}C). No falls or harness catches occurred in any trial. 

\begin{figure}
    \centering
    \includegraphics[width=\linewidth]{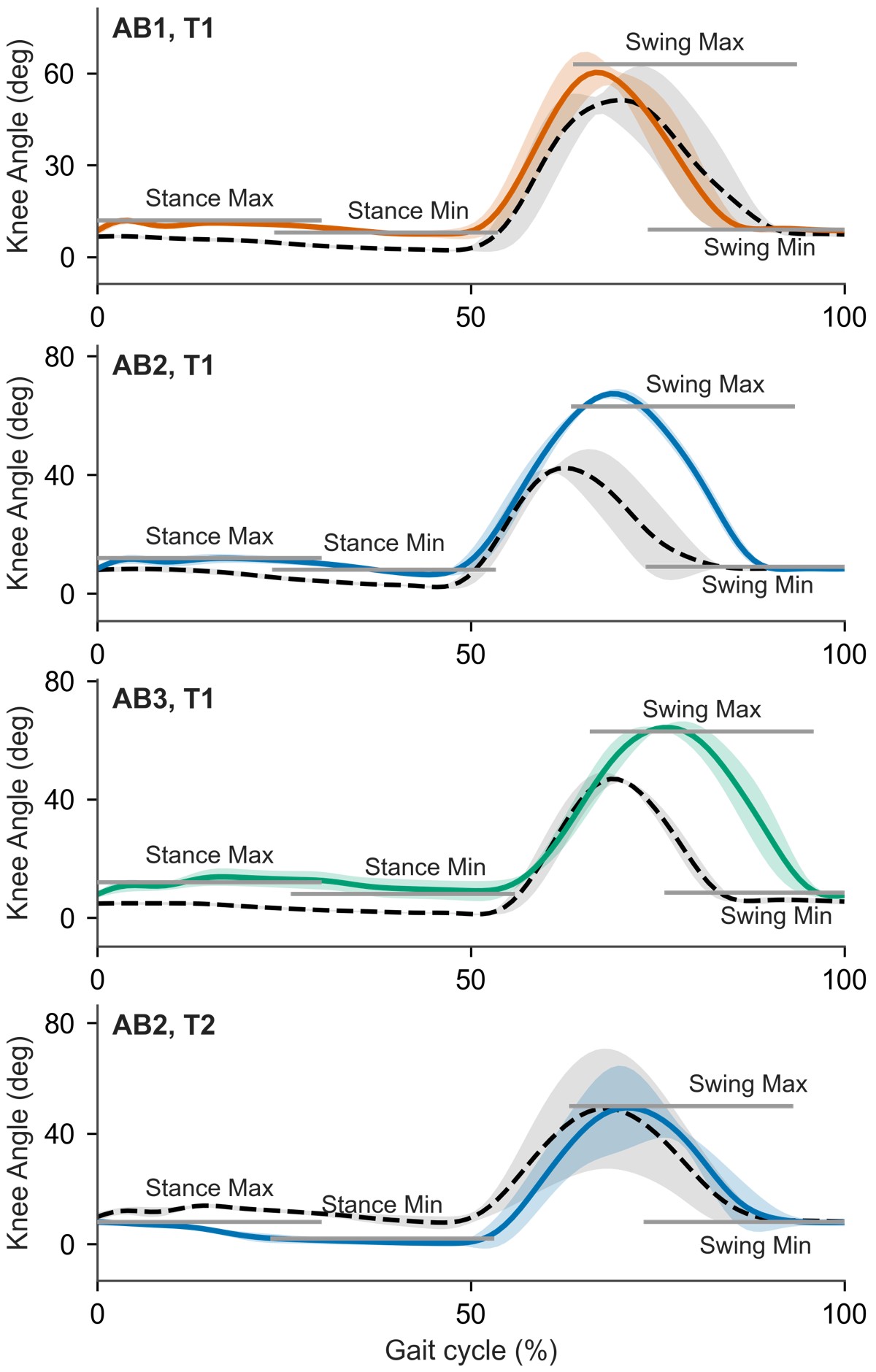}
    \caption{Knee angle over the gait
    cycle at the start of tuning (first four gait cycles, dashed black) and at
    the end (last tuning step, colored), with the target of each feature
    ({gray} bars); shading shows the spread across gait cycles.}
    \label{fig:knee}
\end{figure}


\subsection{Individual Tuning Toward Two Targets}
With target T1, AB1, AB2, and AB3 reached the goal in 8, 10, and 18 RL steps, corresponding to 1.7, 2.3, and 4.2~min of walking (Figs.~\ref{fig:result_main}A and~\ref{fig:stats}B). With target T2, AB2 reached the goal in 8 RL steps (1.7~min). Across the four trials, the negative reward decreased from 1.17--2.70 at the first tuning step to 0.06--0.20 at the last, an improvement of 83--97\% (Fig.~\ref{fig:stats}A).

Swing maximum flexion had the largest initial error in the T1 trials, 17.2--21.0~deg, and was reduced to 1.3--4.3~deg by the end of tuning (Fig.~\ref{fig:stats}C). At the end of tuning, all four feature errors were within their goal thresholds in every trial, and the knee-angle profile matched the target in both stance and swing (Figs.~\ref{fig:stats}C and~\ref{fig:knee}).

The tuned profile followed the prescribed target. For AB2, peak swing flexion at the end of tuning was 67.3~deg under T1 and 49.5~deg under T2, close to the respective targets of 63.0 and 50.0~deg (Fig.~\ref{fig:knee}).


\subsection{Tuning Behavior}

Only one of the four trials, AB3 with T1, needed more than $N_b = 12$ RL steps and thus triggered an online policy update (Fig.~\ref{fig:result_main}{A}). To test whether this update improved the policy, we compared \rev{the RMS TD error (\cref{eq:td}) of} the offline and updated policies on data from steps 13--18, which neither policy \rev{had} been exposed to during training. 
The RMS TD error \rev{decreased} by 47\% \rev{with} the updated policy ({\cref{fig:result_main}B}), indicating that online tuning was able to adapt and tune the prosthesis to AB3's unique gait behavior better than the offline policy. However, since this result is based on only six samples, further validation is needed.

{Convergence was not monotonic (\cref{fig:result_main}A, C). In AB2's T1 and T2 trials, stumbles at steps~2 and~3 raised the stance-maximum error to about 7 and 14~deg, respectively, and lowered the reward to about $-2$, but both recovered within one step. In the T1 trial, the swing-maximum error also grew again at steps 6--7, likely due to step-to-step gait variation. Both trials still converged at steps~10 and~8\rev{, respectively, suggesting} that the tuner was robust to stumbles and gait variability.}


\section{Discussion \& Conclusion }
\label{sec:discussion}

This study investigated whether PICE, an RL-based tuning approach previously validated during treadmill walking~\cite{li_toward_2022}, can tune a commercial powered knee prosthesis under simulated clinical conditions. In the clinic, users walk overground at a self-selected speed and must frequently stop, turn, and reinitiate walking because of limited walkway space~\cite{reznick_clinical_2025}. Both factors increase gait variability and could interrupt the continuous stream of data that might be necessary for PICE.

This study showed that every \rev{simulated clinical tuning trial} with target T1 met the success criterion within 18 RL steps, corresponding to less than 5~minutes of walking ({Fig.~\ref{fig:stats}B}), using only the four parameters a prosthetist already adjusts and the knee's onboard sensors, without access to the proprietary controller or laboratory instruments. This indicates that the tuning algorithm remains effective when gait is less consistent than on a treadmill. Because users might have {preferences for the} target knee profile~\cite{alili2023novel}, it is also important that the tuner can drive the knee toward a different target. In our preliminary test with target T2, AB2 reached the goal in 8 RL steps (about 1.7~min), ending with a knee profile clearly different from that obtained under T1 (Fig.~\ref{fig:knee}).

By design, the starting point of tuning was based on a policy learned from a single participant (AB0) walking on a treadmill. \rev{Three of the four trials reached the goal before the first scheduled update ($N_b = 12$; Fig.~\ref{fig:result_main}A), so tuning succeeded with AB0's offline policy alone.} {Only one trial (AB3 with T1) required an online policy update, and it met the success criterion six steps later.} One \rev{possible} reason for the successful convergence from the existing policy is that the participants walked similarly enough for one person's tuning experience to transfer to others, which may not hold across a broader population. Alternatively, the prior may capture an inherent relationship between the clinical parameters and the resulting knee kinematics that is largely independent of the user. Further studies with a larger and more diverse set of participants, including individuals with transfemoral amputation, are needed to determine which of these explanations accounts for the observed transferability of the policy. 

In all conditions, the tuned knee profile reached the target of each feature, resulting in a profile that differed noticeably from the initial profile (Fig.~\ref{fig:knee}). Some profiles showed higher variability in swing than in stance. While some natural variation is expected in overground walking, the swing variability was markedly higher for AB2 with target T2. This may have been caused by the selected target. T2 resulted in less knee flexion than T1 in swing, which may have felt like less toe clearance to the participant and led to less consistent swing. A systematic analysis of how the choice of target profile affects gait should be performed in future work.

This study has several limitations. First, participants were non-disabled and walked with a bypass adapter.  This allowed us to test {the feasibility of prosthesis personalization} under clinically realistic walking conditions before involving the target population, but individuals with transfemoral amputation differ in residual-limb control, socket interface, and gait, so these experiments must be repeated with individuals with transfemoral amputation. Second, the sample is small, with {one to three} participants per condition, so results are reported per participant and no statistical claims are made. Third,  success was defined by reaching experimenter-set knee-kinematic targets, not by clinical benefit or user preference. Fourth, each condition was tuned once, mostly within a single session, so the  repeatability of the tuned parameters across days was not tested. 
Future studies with a sizable cohort of people with transfemoral amputation {should} systematically investigate 1) how the choice of prior policy affects the time to converge, 2) which features to choose to better reflect clinical benefits, 3) which target profiles are appropriate for individual users, and 4) whether these results translate to other terrains\rev{, including ramps and stairs}.

In conclusion, this preliminary study showed that PICE can tune a commercial powered knee prosthesis during self-paced overground walking with frequent stops and turns, reaching the prescribed gait feature targets in under 5~minutes of walking. An offline policy learned from a single participant on a treadmill was sufficient to tune the knee for new participants without online policy updates in most trials. These findings are a first step toward automated, standardized prosthesis tuning in clinical practice.

\section*{Acknowledgment}
\ifblind\else

\rev{The authors thank \"{O}ssur hf. for providing the prosthetic hardware and the participants for their time and effort in making this study possible.}
\fi

\bibliographystyle{IEEEtran}
\bibliography{references}

@article{hafner_physical_2015,
	title = {Physical performance and self-report outcomes associated with use of passive, adaptive, and active prosthetic knees in persons with unilateral, transfemoral amputation: {Randomized} crossover trial},
	volume = {52},
	issn = {0748-7711, 1938-1352},
	shorttitle = {Physical performance and self-report outcomes associated with use of passive, adaptive, and active prosthetic knees in persons with unilateral, transfemoral amputation},
	doi = {10.1682/JRRD.2014.09.0210},
	language = {en},
	number = {6},
	urldate = {2026-11-03},
	journal = {Journal of Rehabilitation Research and Development},
	author = {Hafner, Brian J. and Askew, Robert L.},
	year = {2015},
	pages = {677--700},
	
}

@article{ichimura_spatiotemporal_2022,
	title = {Spatiotemporal gait patterns in individuals with unilateral transfemoral amputation: {A} hierarchical cluster analysis},
	volume = {17},
	issn = {1932-6203},
	shorttitle = {Spatiotemporal gait patterns in individuals with unilateral transfemoral amputation},
	doi = {10.1371/journal.pone.0279593},
	language = {en},
	number = {12},
	urldate = {2026-11-03},
	journal = {PLOS ONE},
	publisher = {Public Library of Science},
	author = {Ichimura, Daisuke and Amma, Ryo and Hisano, Genki and Murata, Hiroto and Hobara, Hiroaki},
	month = dec,
	year = {2022},
	pages = {e0279593},

}

@article{kestur_comparing_2024,
	title = {Comparing the lower limb joint biomechanics of the {Power} {Knee}, {C}-{Leg} and {Rheo} {Knee} during ramp and stair ambulation},
	volume = {171},
	issn = {00219290},
	doi = {10.1016/j.jbiomech.2024.112201},
	language = {en},
	urldate = {2026-11-03},
	journal = {Journal of Biomechanics},
	author = {Kestur, Sujay and Zhou, Sixu and O’Sullivan, Gwyn and Young, Aaron and Herrin, Kinsey},
	month = jun,
	year = {2024},
	pages = {112201},
}

@article{herrin_toward_2026,
	title = {Toward personalizing prosthesis prescription: {A} take‐home study of three microprocessor‐controlled prosthetic knees: {A} randomized crossover study},
	volume = {18},
	issn = {1934-1482, 1934-1563},
	shorttitle = {Toward personalizing prosthesis prescription},
	doi = {10.1002/pmrj.70028},
	language = {en},
	number = {3},
	urldate = {2026-11-03},
	journal = {PM\&R},
	author = {Herrin, Kinsey and Kestur, Sujay and Zhou, Sixu and O'Sullivan, Gwyn and Snow, Teresa and Childers, Walter Lee and Young, Aaron},
	month = mar,
	year = {2026},
	pages = {288--304},
}

@article{wen_online_2020,
	title = {Online {Reinforcement} {Learning} {Control} for the {Personalization} of a {Robotic} {Knee} {Prosthesis}},
	volume = {50},
	copyright = {https://ieeexplore.ieee.org/Xplorehelp/downloads/license-information/IEEE.html},
	issn = {2168-2267, 2168-2275},
	doi = {10.1109/TCYB.2019.2890974},
	language = {en},
	number = {6},
	journal = {IEEE Transactions on Cybernetics},
	author = {Wen, Yue and Si, Jennie and Brandt, Andrea and Gao, Xiang and Huang, He Helen},
	month = jun,
	year = {2020},
	pages = {2346--2356},
}

@article{huang_taking_2021,
	title = {Taking both sides: seeking symbiosis between intelligent prostheses and human motor control during locomotion},
	volume = {20},
	issn = {24684511},
	shorttitle = {Taking both sides},
	doi = {10.1016/j.cobme.2021.100314},
	language = {en},
	urldate = {2025-05-12},
	journal = {Current Opinion in Biomedical Engineering},
	author = {Huang, He (Helen) and Si, Jennie and Brandt, Andrea and Li, Minhan},
	month = dec,
	year = {2021},
	pages = {100314},
}

@article{li_toward_2022,
	title = {Toward {Expedited} {Impedance} {Tuning} of a {Robotic} {Prosthesis} for {Personalized} {Gait} {Assistance} by {Reinforcement} {Learning} {Control}},
	volume = {38},
	copyright = {https://ieeexplore.ieee.org/Xplorehelp/downloads/license-information/IEEE.html},
	issn = {1552-3098, 1941-0468},
	doi = {10.1109/TRO.2021.3078317},
	language = {en},
	number = {1},
	urldate = {2025-05-12},
	journal = {IEEE Transactions on Robotics},
	author = {Li, Minhan and Wen, Yue and Gao, Xiang and Si, Jennie and Huang, He},
	month = feb,
	year = {2022},
	pages = {407--420},
}

@article{rogers-bradley_variable-stiffness_2024,
	title = {Variable-stiffness prosthesis improves biomechanics of walking across speeds compared to a passive device},
	volume = {14},
	copyright = {2024 The Author(s)},
	issn = {2045-2322},
	doi = {10.1038/s41598-024-67230-3},
	language = {en},
	number = {1},
	urldate = {2025-09-22},
	journal = {Scientific Reports},
	publisher = {Nature Publishing Group},
	author = {Rogers-Bradley, Emily and Yeon, Seong Ho and Landis, Christian and Lee, Duncan R. C. and Herr, Hugh M.},
	month = jul,
	year = {2024},
	pages = {16521},
}

@article{lawson_robotic_2014,
	title = {A {Robotic} {Leg} {Prosthesis}: {Design}, {Control}, and {Implementation}},
	volume = {21},
	copyright = {https://ieeexplore.ieee.org/Xplorehelp/downloads/license-information/IEEE.html},
	issn = {1070-9932, 1558-223X},
	shorttitle = {A {Robotic} {Leg} {Prosthesis}},
	doi = {10.1109/MRA.2014.2360303},
	language = {en},
	number = {4},
	urldate = {2026-11-03},
	journal = {IEEE Robotics \& Automation Magazine},
	author = {Lawson, Brian E. and Mitchell, Jason and Truex, Don and Shultz, Amanda and Ledoux, Elissa and Goldfarb, Michael},
	month = dec,
	year = {2014},
	pages = {70--81},
}

@article{reznick_clinical_2025,
	title = {A Clinical Tuning Framework for Continuous Kinematic and Impedance Control of a Powered Knee-Ankle Prosthesis},
	author = {Reznick, Emma and Best, T. Kevin and Gregg, Robert D.},
	journal = {IEEE Journal of Translational Engineering in Health and Medicine},
	volume = {13},
	pages = {227--236},
	year = {2025},
	doi = {10.1109/JTEHM.2025.3567578}
}

@article{zhao_first_2017,
	title = {First steps toward translating robotic walking to prostheses: a nonlinear optimization based control approach},
	author = {Zhao, Huihua and Horn, Jonathan and Reher, Jacob and Paredes, Victor and Ames, Aaron D.},
	journal = {Autonomous Robots},
	volume = {41},
	number = {3},
	pages = {725--742},
	year = {2017},
	doi = {10.1007/s10514-016-9565-1}
}

@article{gehlhar_review_2023,
	title = {A review of current state-of-the-art control methods for lower-limb powered prostheses},
	volume = {55},
	issn = {13675788},
	doi = {10.1016/j.arcontrol.2023.03.003},
	language = {en},
	urldate = {2026-06-21},
	journal = {Annual Reviews in Control},
	author = {Gehlhar, Rachel and Tucker, Maegan and Young, Aaron J. and Ames, Aaron D.},
	year = {2023},
	pages = {142--164},
}

@misc{le_replay-constrained_2026,
	title = {A {Replay}-{Constrained} {Simulation} {Framework} for {Personalization} of {Powered} {Knee}--{Ankle} {Prosthesis} {Controllers}},
	doi = {10.48550/arXiv.2607.22858},
	urldate = {2026-08-04},
	publisher = {arXiv},
	author = {Le, Duong and Posh, Ryan and Cheng, Shihao and Ghaffari, Maani and Gregg, Robert D.},
	month = jul,
	year = {2026},
	note = {arXiv:2607.22858 [cs.RO]},
}

@article{best_clinical_2025,
	title = {The clinical effects of the Össur {Power} {Knee} with phase-based and default control during sitting, standing, and walking},
	volume = {22},
	issn = {1743-0003},
	doi = {10.1186/s12984-025-01729-2},
	urldate = {2026-09-07},
	journal = {Journal of NeuroEngineering and Rehabilitation},
	author = {Best, T. Kevin and Seelhoff, C. Andrew and Wensman, Jeffrey and Gregg, Robert D.},
	month = sep,
	year = {2025},
	pages = {200},
}

@inproceedings{hong_offline_2025,
	address = {Chicago, IL, USA},
	title = {An {Offline} {Reinforcement} {Learning}-{Based} {Auto}-{Tuning} {Framework} for {Continuous} {Impedance} {Control} in {Powered} {Prostheses}},
	copyright = {https://doi.org/10.15223/policy-029},
	isbn = {979-8-3503-8068-2},
	doi = {10.1109/ICORR66766.2025.11063218},
	language = {en},
	urldate = {2026-10-05},
	booktitle = {2025 {International} {Conference} on {Rehabilitation} {Robotics} ({ICORR})},
	publisher = {IEEE},
	author = {Hong, Woolim and Alili, Abbas and Huang, He Helen},
	month = may,
	year = {2025},
	pages = {1829--1834},
}

@article{slade_human---loop_2024,
	title = {On human-in-the-loop optimization of human–robot interaction},
	volume = {633},
	copyright = {2024 Springer Nature Limited},
	issn = {1476-4687},
	doi = {10.1038/s41586-024-07697-2},
	language = {en},
	number = {8031},
	urldate = {2026-10-05},
	journal = {Nature},
	publisher = {Nature Publishing Group},
	author = {Slade, Patrick and Atkeson, Christopher and Donelan, J. Maxwell and Houdijk, Han and Ingraham, Kimberly A. and Kim, Myunghee and Kong, Kyoungchul and Poggensee, Katherine L. and Riener, Robert and Steinert, Martin and Zhang, Juanjuan and Collins, Steven H.},
	month = sep,
	year = {2024},
	pages = {779--788},
}

@article{simon_configuring_2014,
	title = {Configuring a {Powered} {Knee} and {Ankle} {Prosthesis} for {Transfemoral} {Amputees} within {Five} {Specific} {Ambulation} {Modes}},
	volume = {9},
	issn = {1932-6203},
	doi = {10.1371/journal.pone.0099387},
	language = {en},
	number = {6},
	urldate = {2026-10-08},
	journal = {PLOS ONE},
	publisher = {Public Library of Science},
	author = {Simon, Ann M. and Ingraham, Kimberly A. and Fey, Nicholas P. and Finucane, Suzanne B. and Lipschutz, Robert D. and Young, Aaron J. and Hargrove, Levi J.},
	month = jun,
	year = {2014},
	pages = {e99387},
}

@article{tucker_control_2015,
	title = {Control strategies for active lower extremity prosthetics and orthotics: a review},
	volume = {12},
	issn = {1743-0003},
	shorttitle = {Control strategies for active lower extremity prosthetics and orthotics},
	doi = {10.1186/1743-0003-12-1},
	language = {en},
	number = {1},
	urldate = {2026-10-08},
	journal = {Journal of NeuroEngineering and Rehabilitation},
	author = {Tucker, Michael R. and Olivier, Jeremy and Pagel, Anna and Bleuler, Hannes and Bouri, Mohamed and Lambercy, Olivier and Millán, José del R. and Riener, Robert and Vallery, Heike and Gassert, Roger},
	month = jan,
	year = {2015},
	pages = {1},
	
}

@article{feldman_estimating_2026,
	title = {Estimating and interpreting how humans prioritize multiple movement goals during walking},
	volume = {23},
	issn = {1743-0003},
	doi = {10.1186/s12984-026-01972-1},
	language = {en},
	number = {1},
	urldate = {2026-10-11},
	journal = {Journal of NeuroEngineering and Rehabilitation},
	author = {Feldman, Jordan N. and Morales Loro, David and Chin, Andrew and Erickson, Sarah and Li, Sarah and Bonbrest, Eliana and Slade, Patrick},
	month = apr,
	year = {2026},
	pages = {184},
}

@article{kulkarni_humans_2023,
	title = {Humans prioritize walking efficiency or walking stability based on environmental risk},
	volume = {18},
	issn = {1932-6203},
	doi = {10.1371/journal.pone.0284278},
	language = {en},
	number = {4},
	urldate = {2026-10-11},
	journal = {PLOS ONE},
	publisher = {Public Library of Science},
	author = {Kulkarni, Ashwini and Cui, Chuyi and Rietdyk, Shirley and Ambike, Satyajit},
	month = apr,
	year = {2023},
	pages = {e0284278},
}

@article{zhang_human---loop_2017,
	title = {Human-in-the-loop optimization of exoskeleton assistance during walking},
	volume = {356},
	doi = {10.1126/science.aal5054},
	number = {6344},
	urldate = {2026-10-26},
	journal = {Science},
	publisher = {American Association for the Advancement of Science},
	author = {Zhang, Juanjuan and Fiers, Pieter and Witte, Kirby A. and Jackson, Rachel W. and Poggensee, Katherine L. and Atkeson, Christopher G. and Collins, Steven H.},
	month = jun,
	year = {2017},
	pages = {1280--1284},
}

@article{ding_human---loop_2018,
	title = {Human-in-the-loop optimization of hip assistance with a soft exosuit during walking},
	volume = {3},
	doi = {10.1126/scirobotics.aar5438},
	number = {15},
	urldate = {2026-10-26},
	journal = {Science Robotics},
	publisher = {American Association for the Advancement of Science},
	author = {Ding, Ye and Kim, Myunghee and Kuindersma, Scott and Walsh, Conor J.},
	month = feb,
	year = {2018},
	pages = {eaar5438},
}

@article{welker_shortcomings_nodate,
	title = {Shortcomings of human-in-the-loop optimization of an ankle-foot prosthesis emulator: a case series},
	volume = {8},
	issn = {2054-5703},
	shorttitle = {Shortcomings of human-in-the-loop optimization of an ankle-foot prosthesis emulator},
	doi = {10.1098/rsos.202020},
	number = {5},
	urldate = {2026-10-26},
    year = {2021}, 
	journal = {Royal Society Open Science},
	author = {Welker, Cara Gonzalez and Voloshina, Alexandra S. and Chiu, Vincent L. and Collins, Steven H.},
	pages = {202020},
}

@article{fylstra_human-prosthesis_2022,
  title = {Human-prosthesis cooperation: combining adaptive prosthesis control with visual feedback guided gait},
  volume = {19},
  issn = {1743-0003},
  shorttitle = {Human-prosthesis cooperation},
  doi = {10.1186/s12984-022-01118-z},
  language = {en},
  number = {1},
  urldate = {2025-05-12},
  journal = {Journal of NeuroEngineering and Rehabilitation},
  author = {Fylstra, Bretta L. and Lee, I-Chieh and Li, Minhan and Lewek, Michael D. and Huang, He},
  month = dec,
  year = {2022},
  pages = {140},
}

@article{liu_addressing_2026,
	title = {Addressing {Human}–{Robot} {Symbiosis} via {Bilevel} {Optimization} of {Robotic} {Knee} {Prosthesis} {Control}},
	volume = {42},
	issn = {1941-0468},
	doi = {10.1109/TRO.2025.3634368},
	urldate = {2026-10-28},
	journal = {IEEE Transactions on Robotics},
	author = {Liu, Wentao and Nalam, Varun and Si, Jennie and Huang, He},
	year = {2026},
	pages = {170--183},
}

@article{lagoudakis_least-squares_2003,
	title = {Least-{Squares} {Policy} {Iteration}},
	volume = {4},
	issn = {ISSN 1533-7928},
	month = dec,
	urldate = {2026-10-29},
	journal = {Journal of Machine Learning Research},
	author = {Lagoudakis, Michail G. and Parr, Ronald},
	year = {2003},
	pages = {1107--1149},
}

@article{hollman2016comparison,
  title={A comparison of variability in spatiotemporal gait parameters between treadmill and overground walking conditions},
  author={Hollman, John H and Watkins, Molly K and Imhoff, Angela C and Braun, Carly E and Akervik, Kristen A and Ness, Debra K},
  journal={Gait \& Posture},
  volume={43},
  pages={204--209},
  year={2016},
  publisher={Elsevier}
}

@article{alili2023novel,
  title={A novel framework to facilitate user preferred tuning for a robotic knee prosthesis},
  author={Alili, Abbas and Nalam, Varun and Li, Minhan and Liu, Ming and Feng, Jing and Si, Jennie and Huang, He},
  journal={IEEE Transactions on Neural Systems and Rehabilitation Engineering},
  volume={31},
  pages={895--903},
  year={2023},
  publisher={IEEE}
}

\end{document}